\documentclass[sigconf]{acmart}
\usepackage{mathtools}
\DeclarePairedDelimiter\ceil{\lceil}{\rceil}
\usepackage{booktabs} 
\usepackage{tikz}
\usepackage{array}
\usetikzlibrary{positioning, fit, arrows.meta}
\usepackage[group-separator={,},
            group-four-digits,
            output-decimal-marker={.}]{siunitx}

\setcopyright{rightsretained}

\setcopyright{rightsretained}
\acmConference[SIGIR 2018 eCom]{ACM SIGIR Workshop on eCommerce}{July 2018}{Ann Arbor, Michigan, USA} 
\acmYear{2018}
\copyrightyear{2018}

\begin{document}

\title[Predicting purchasing intent: Automatic Feature Learning using RNNs]{Predicting purchasing intent: Automatic Feature Learning using Recurrent Neural Networks}

\author{Humphrey Sheil}
\affiliation{%
  \institution{Cardiff University}
  \streetaddress{5 The Parade, Roath}
  \city{Cardiff}
  \state{Wales}
  \postcode{CF24 3AA}
}
\email{sheilh@cardiff.ac.uk}

\author{Omer Rana}
\affiliation{%
  \institution{Cardiff University}
  \streetaddress{5 The Parade, Roath}
  \city{Cardiff}
  \state{Wales}
  \postcode{CF24 3AA}
}
\email{RanaOF@cardiff.ac.uk}

\author{Ronan Reilly}
\affiliation{%
  \institution{Maynooth University}
  \city{Maynooth}
  \state{Ireland}
}
\email{Ronan.Reilly@mu.ie}

\renewcommand{\shortauthors}{H. Sheil et al.}

\begin{abstract}
We present a neural network for predicting purchasing intent in an Ecommerce setting. Our main contribution is to address the significant investment in feature engineering that is usually associated with state-of-the-art methods such as Gradient Boosted Machines. We use trainable vector spaces to model varied, semi-structured input data comprising categoricals, quantities and unique instances. Multi-layer recurrent neural networks capture both session-local and dataset-global event dependencies and relationships for user sessions of any length. An exploration of model design decisions including parameter sharing and skip connections further increase model accuracy. Results on benchmark datasets deliver classification accuracy within 98\% of state-of-the-art on one and exceed state-of-the-art on the second without the need for any domain / dataset-specific feature engineering on both short and long event sequences.
\end{abstract}

%
%


\maketitle

\section{Introduction}

In the Ecommerce domain, merchants can increase their sales volume and profit margin by acquiring better answers for two questions:
\begin{itemize}
\item Which users are most likely to purchase (predict purchasing intent).
\item Which elements of the product catalogue do users prefer (rank content).
\end{itemize}

By how much can merchants realistically increase profits? Table \ref{table:profit} illustrates that merchants can improve profit by between 2\% and 11\% depending on the contributing variable. In the fluid and highly competitive world of online retailing, these margins are significant, and understanding a user's shopping intent can positively influence three out of four major variables that affect profit. In addition, merchants increasingly rely on (and pay advertising to) much larger third-party portals (for example eBay, Google, Bing, Taobao, Amazon) to achieve their distribution, so any direct measures the merchant group can use to increase their profit is sorely needed.

\begin{table}[!htb]
\centering
\begin{tabular}{ |c|c|c|c| } 
 \hline
  & McKinsey & A.T. Kearney & Affected by \\
  &&& shopping intent\\
  \hline
Price management & 11.1\% & 8.2\% & Yes \\ 
 \hline
Variable cost & 7.8\% & 5.1\% & Yes \\ 
 \hline
Sales volume & 3.3\% & 3.0\% & Yes \\ 
 \hline
Fixed cost & 2.3\% & 2.0\% & No \\ 
 \hline
\end{tabular}
\captionof{table}{Effect of improving different variables on operating profit, from \cite{phillips2005pricing}. In three out of four categories, knowing more about a user's shopping intent can be used to improve merchant profit.}
\label{table:profit}
\end{table}

Virtually all Ecommerce systems can be thought of as a generator of clickstream data - a log of \{item - userid - action\} tuples which captures user interactions with the system. A chronological grouping of these tuples by user ID is commonly known as a \emph{session}.

Predicting a users intent to purchase is more difficult than ranking content for the following reasons \cite{Sheil2017}: Clickers (users who only click and never purchase within a session) and buyers (users who click and also purchase at least one item within a single session) can appear to be very similar, right up until a purchase action occurs. Additionally, the ratio between clickers and buyers is always heavily imbalanced - and can be 20:1 in favour of clickers or higher. An uninterested user will often click on an item during browsing as there is no cost to doing so - an uninterested user will not \emph{purchase} an item however. In our opinion, this user behaviour is in stark contrast to other settings such as predicting if a user will "like" or "pin" a piece of content hosted on a social media platform after viewing it, where there is no monetary amount at stake for the user. As noted in \cite{toth2017predicting}, shoppers behave differently when visiting online vs physical stores and online conversion rates are substantially lower, for a variety of reasons.

When a merchant has increased confidence that a subset of users are more likely to purchase, they can use this information in the form of proactive actions to maximize conversion and yield. The merchant may offer a time-limited discount, spend more on targeted (and relevant) advertising to re-engage these users, create bundles of complementary products to push the user to complete their purchase, or even offer a lower-priced own-brand alternative if the product is deemed to be fungible.

However there are counterweights to the desire to create more and more accurate models of online user behaviour - namely user privacy and ease of implementation. Users are increasingly reluctant to share personal information with online services, while complex Machine Learning models are difficult to implement and maintain in a production setting \cite{high-interest}.

We surveyed existing work in this area and found that well-performing approaches have a number of factors in common:

\begin{itemize}
\item Heavy investment in dataset-specific feature engineering was necessary, regardless of the model implementation chosen.
\item Model choices favour techniques such as Gradient Boosted Machines (GBM) \cite{Friedman00greedyfunction} and Field-aware Factorization Machines (FFM) \cite{Rendle:2010:FM:1933307.1934620} which are well-suited to creating representations of semi-structured clickstream data once good features have been developed \cite{Romov:2015:RCE:2813448.2813510}, \cite{Yan:2015:EIR:2813448.2813511}, \cite{Volkovs:2015:TAI:2813448.2813512}.
\end{itemize}

In \cite{Sheil2017}, an important feature class employed the notion of item similarity, modelled as a learned vector space generated by word2vec \cite{DBLP:journals/corr/abs-1301-3781} and calculated using a standard pairwise cosine metric between item vectors. In an Ecommerce context, items are more similar if they co-occur frequently over all user sessions in the corpus and are dissimilar if they infrequently co-occur. The items themselves may be physically dissimilar (for example - headphones and batteries), but they are often browsed and purchased together.

However, in common with other work, \cite{Sheil2017} still requires a heavy investment in feature engineering. The drawback of specific features is how tied they are to either a domain, dataset or both. The ability of Deep Learning to discover good representations without explicit feature engineering is well-known \cite{Goodfellow-et-al-2016}. In addition, Artificial neural networks (ANNs) perform well with distributed representations such as embeddings, and ANNs with a recurrence capability to model events over time - Recurrent Neural Networks (RNNs) - are well-suited to sequence processing and labelling tasks \cite{DBLP:journals/corr/Lipton15}.

Our motivation then is to build a good model of user intent prediction which does not rely on private user data, and is also straightforward to implement in a real-world environment. What performance can RNNs with an appropriate input representation and end-to-end training regime achieve on the prediction of purchasing intent task? Can this performance be achieved within the constraint of only processing anonymous session data and remaining straightforward to implement on other Ecommerce datasets?

\section{Related Work}
The problem of user intent or session classification in an online setting has been heavily studied, with a variety of classic Machine Learning and Deep Learning modelling techniques employed. \cite{Romov:2015:RCE:2813448.2813510} was the original winner of the competition using one of the the datasets considered here using a commercial implementation of GBM with extensive feature engineering and is still to our knowledge the State of the Art (SOTA) implementation for this dataset. However the paper authors did make their model predictions freely available and we use these in the Experiments section to compare our model performance to theirs.

\cite{DBLP:journals/corr/HidasiKBT15} uses RNNs on a subset of the same dataset to predict the next session click (regardless of user intent) so removed 1-click sessions and merged clickers and buyers, whereas this work remains focused on the user intent classification problem. \cite{DBLP:journals/corr/abs-1803-09587} compares \cite{DBLP:journals/corr/HidasiKBT15} to a variety of classical Machine Learning algorithms on multiple datasets and finds that performance varies considerably by dataset. \cite{Zhu2017WhatTD} extends \cite{DBLP:journals/corr/HidasiKBT15} with a variant of LSTM to capture variations in dwelltime between user actions. User dwelltime is considered an important factor in multiple implementations and has been addressed in multiple ways. For shopping behaviour prediction, \cite{toth2017predicting} uses a mixture of Recurrent Neural Networks and treats the problem as a sequence-to-sequence translation problem, effectively combining two models  (prediction and recommendation) into one. However only sessions of length 4 or greater are considered - removing the bulk from consideration. From \cite{Jannach2017SessionbasedIR}, we know that short sessions are very common in Ecommerce datasets, moreover a user's most recent actions are often more important in deciphering their intent than older actions. Therefore we argue that all session lengths should be included. \cite{Pryzant2017PredictingSF} adopts a tangential approach - still focused on predicting purchases, but using textual product metadata to correlate words and terms that suit a particular geographic market better than others. Broadening our focus to include the general use of RNNs in the Ecommerce domain, Recurrent Recommender Networks are used in \cite{Wu:2017:RRN:3018661.3018689} to incorporate temporal features with user preferences to improve recommendations, to predict future behavioural directions, but not purchase intent. \cite{DBLP:journals/corr/TanXL16} further extends \cite{DBLP:journals/corr/HidasiKBT15} by focusing on data augmentation and compensating for shifts in the underlying distribution of the data.

In \cite{Koren:2009:CFT:1557019.1557072}, the authors augment a more classical Machine Learning approach (Singular Value Decomposition or SVD) to better capture temporal information to predict user behaviour - an alternative approach to the unrolling methodology used in this paper.

Using embeddings as a learned representation is a common technique. In \cite{DBLP:journals/corr/BarkanK16}, embeddings are used to model items in a low dimensional space to calculate a similarity metric, however temporal ordering is discarded. Learnable embeddings are also used in \cite{DBLP:journals/corr/GrbovicRDBSBS16} to model items and used purchase confirmation emails as a high-quality signal of user intent.
Unrolling events that exceed an arbitrary threshold to create a better input representation for user dwelltime or interest is addressed in \cite{Bogina2017IncorporatingDT}. In \cite{Ding:2015:MUC:2886521.2886653}, Convolutional Neural Networks (CNNs) are used as the model implementation and micro-blogging content is analyzed rather than an Ecommerce clickstream.

\section{Our Approach}

Classical Machine Learning approaches such as GBM work well and are widely used on Ecommerce data, at least in part because the data is \emph{structured}. GBM is an efficient model as it enables an additive expansion in a set of basis functions or weak learners to continually minimize a residual error. One weakness of GBM is a propensity for overly-deep or wide decision trees to over-fit the training data and thus record poor performance on the validation and test set due to high variance \cite{Vezhnevets:2007:ABO:1421665.1421707}, \cite{Volkovs:2015:TAI:2813448.2813512}. although this can be controlled using hyperparameters (namely tree depth, learning rate, minimum weight to split a tree node (min\_child\_weight) and data sub-sampling). GBM also requires significant feature engineering effort and does not naturally process the sequence in order, rather it consumes a compressed version of it (although it is possible to provide a one-hot vector representation of the input sequence as a feature).
Our approach is dual in nature - firstly we construct an input representation for clickstream / session data that eliminates the need for feature engineering. Second, we design a model which can consume this input representation and predict user purchase intent in an end-to-end, sequence to prediction manner.

\subsection{Embeddings as item / word representations}

Natural Language Processing (NLP) tasks, such as information retrieval, part-of-speech tagging and chunking, operate by assigning a probability value to a sequence of words. To this end, language models have been developed, defining a mathematical model to capture statistical properties of words and the dependencies among them.

Learning good representations of input data is a central task in designing a machine learning model that can perform well. An \emph{embedding} is a vector space model where words are converted to a low-dimensional vector. Vector space models embed words where semantically similar words are mapped to nearby points. Popular generators of word to vector mappings such as \cite{DBLP:journals/corr/abs-1301-3781} and \cite{pennington2014glove}, operate in an unsupervised manner - predicting similarity or minimizing a perplexity metric using word co-occurrence counts over a target corpus.
We decided to employ embeddings as our target representation since:
\begin{itemize}
\item We can train the embeddings layer at the same time as training the model itself - promoting simplicity.
\item Ecommerce data is straightforward to model as a dictionary of words.
\item Embedding size can be increased or decreased based on dictionary size and word complexity during the architecture tuning / hyper parameter search phase.
\end{itemize}

Unlike \cite{DBLP:journals/corr/abs-1301-3781} and \cite{pennington2014glove}, we chose not to pre-train the embeddings to minimize a perplexity error measure. Instead we allow the model to modify the embedding weights at training time by back-propagating the loss from a binary classification criterion.

\subsection{Recurrent Neural Networks}

Recurrent neural networks \cite{Rumelhart:1986:LIR:104279.104293} (RNNs) are a specialized class of neural networks for processing sequential data.
A recurrent network is deep in \emph{time} rather than space and arranges hidden state vectors $h^l_t$ in a two-dimensional grid, where $t = 1 \ldots T$ is thought of as time and $l = 1 \ldots L$ is the depth.
All intermediate vectors $h^l_t$ are computed as a function of $h^l_{t-1}$ and $h_t^{l{-}1}$. Through these hidden vectors, each output $y$ at some particular time step $t$ becomes an approximating function of all input vectors up to that time, ${x_1 ,\ldots, x_t}$ \cite{DBLP:journals/corr/KarpathyJL15}.

\subsubsection{LSTM and GRU}
Long Short-Term Memory (LSTM) \cite{Hochreiter:1997:LSM:1246443.1246450} is an extension to colloquial or vanilla RNNs designed to address the twin problems of vanishing and exploding gradients during training \cite{Pascanu:2013:DTR:3042817.3043083}. Vanishing gradients make learning difficult as the correct (downward) trajectory of the gradient is difficult to discern, while exploding gradients make training unstable - both are undesirable outcomes. Long-term dependencies in the input data, causing a deep computational graph which must iterate over the data are the root cause of vanishing / exploding gradients. \cite{Goodfellow-et-al-2016} explain this phenomenon succinctly. Like all deep learning models, RNNs require multiplication by a matrix $W$. After $t$ steps, this equates to multiplying by $W^t$. Therefore:

\begin{equation}
    \label{eqn:gradients}
W^t = (Vdiag(\lambda)V^{-1})^t = Vdiag(\lambda)^tV^{-1}
\end{equation}

Eigenvalues ($\lambda$) that are not more or less equal to 1 will either explode if they are $> 1$, or vanish if they are $< 1$. Gradients will then be scaled by $diag(\lambda)^t$.

LSTM solves this problem by possessing an internal recurrence, which stabilizes the gradient flow, even over long sequences. However this comes at a price of complexity. For each element in the input sequence, each layer computes the following function:

\begin{equation}
\begin{split}\begin{array}{ll}
i_t = \sigma(W_{ii} x_t + b_{ii} + W_{hi} h_{(t-1)} + b_{hi}) \\
f_t = \sigma(W_{if} x_t + b_{if} + W_{hf} h_{(t-1)} + b_{hf}) \\
g_t = \tanh(W_{ig} x_t + b_{ig} + W_{hc} h_{(t-1)} + b_{hg}) \\
o_t = \sigma(W_{io} x_t + b_{io} + W_{ho} h_{(t-1)} + b_{ho}) \\
c_t = f_t * c_{(t-1)} + i_t * g_t \\
h_t = o_t * \tanh(c_t)
\end{array}\end{split}
\end{equation}

where:

$h_t$ is the hidden state at time t,\\
$c_t$ is the cell state at time t,\\
$x_t$ is the hidden state of the previous layer at time $t$ or $input_t$ for the first layer,\\
$i_t$, $f_t$, $g_t$, $o_t$ are the input, forget, cell, and out gates, respectively,\\
$\sigma$ is the sigmoid function.

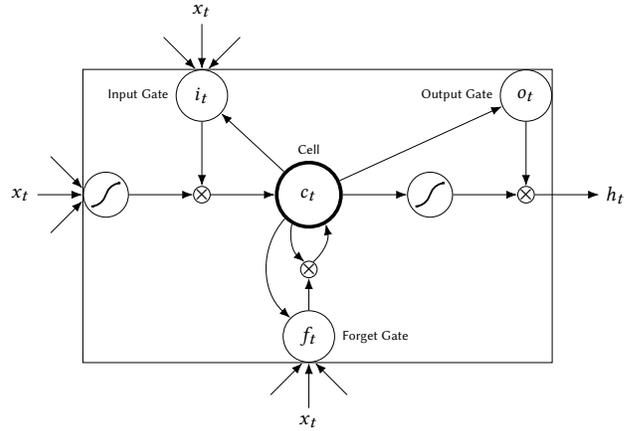
\begin{figure}
 \centering
\resizebox {\columnwidth} {!} {
\begin{tikzpicture}[
    prod/.style={circle, draw, inner sep=0pt},
    ct/.style={circle, draw, inner sep=5pt, ultra thick, minimum width=10mm},
    ft/.style={circle, draw, minimum width=8mm, inner sep=1pt},
    filter/.style={circle, draw, minimum width=7mm, inner sep=1pt, path picture={\draw[thick, rounded corners] (path picture bounding box.center)--++(65:2mm)--++(0:1mm);
    \draw[thick, rounded corners] (path picture bounding box.center)--++(245:2mm)--++(180:1mm);}},
    mylabel/.style={font=\scriptsize\sffamily},
    >=LaTeX
    ]

\node[ct, label={[mylabel]Cell}] (ct) {$c_t$};
\node[filter, right=of ct] (int1) {};
\node[prod, right=of int1] (x1) {$\times$}; 
\node[right=of x1] (ht) {$h_t$};
\node[prod, left=of ct] (x2) {$\times$}; 
\node[filter, left=of x2] (int2) {};
\node[prod, below=5mm of ct] (x3) {$\times$}; 
\node[ft, below=5mm of x3, label={[mylabel]right:Forget Gate}] (ft) {$f_t$};
\node[ft, above=of x2, label={[mylabel]left:Input Gate}] (it) {$i_t$};
\node[ft, above=of x1, label={[mylabel]left:Output Gate}] (ot) {$o_t$};

\foreach \i/\j in {int2/x2, x2/ct, ct/int1, int1/x1,
            x1/ht, it/x2, ct/it, ct/ot, ot/x1, ft/x3}
    \draw[->] (\i)--(\j);

\draw[->] (ct) to[bend right=45] (ft);

\draw[->] (ct) to[bend right=30] (x3);
\draw[->] (x3) to[bend right=30] (ct);

\node[fit=(int2) (it) (ot) (ft), draw, inner sep=0pt] (fit) {};

\draw[<-] (fit.west|-int2) coordinate (aux)--++(180:7mm) node[left]{$x_t$};
\draw[<-] ([yshift=1mm]aux)--++(135:7mm);
\draw[<-] ([yshift=-1mm]aux)--++(-135:7mm);

\draw[<-] (fit.north-|it) coordinate (aux)--++(90:7mm) node[above]{$x_t$};
\draw[<-] ([xshift=1mm]aux)--++(45:7mm);
\draw[<-] ([xshift=-1mm]aux)--++(135:7mm);


\draw[<-] (fit.south-|ft) coordinate (aux)--++(-90:7mm) node[below]{$x_t$};
\draw[<-] ([xshift=1mm]aux)--++(-45:7mm);
\draw[<-] ([xshift=-1mm]aux)--++(-135:7mm);
\end{tikzpicture}
}
\caption{A single LSTM cell, depicting the hidden and cell states, as well as the three gates controlling memory (input, forget and output).} \label{fig:lstm}
\end{figure}

Gated Recurrent Units, or GRU \cite{DBLP:journals/corr/ChoMBB14} are a simplification of LSTM, with one less gate and the hidden state and cell state vectors combined. In practice, both LSTM and GRU are used interchangeably and the performance difference between both cell types is often minimal and / or dataset-specific.

\section{Implementation}

\subsection{Datasets used}

The RecSys 2015 Challenge \cite{Ben-Shimon:2015:RCY:2792838.2798723} and the Retail Rocket Kaggle \cite{retailrocket} datasets provide anonymous Ecommerce clickstream data well suited to testing purchase prediction models.
Both datasets are reasonable in size - consisting of 9.2 million and 1.4 million user sessions respectively. These sessions are anonymous and consist of a  chronological sequence of time-stamped events describing user interactions (clicks) with content while browsing and shopping online. The logic used to mark the start and end of a user session is dataset-specific - the RecSys 2015 dataset contains more sessions with a small item catalogue while the Retail Rocket dataset contains less sessions with an item catalogue ~5x larger than the RecSys 2015 dataset. Both datasets contain a very high proportion of short length sessions ($<= 3$ events), making this problem setting quite difficult for RNNs to solve. The Retail Rocket dataset contains much longer sessions when measured by duration - the RecSys 2015 user sessions are much shorter in duration. In summary, the datasets differ in important respects, and provide a good test of model generalization ability.

For both datasets, no sessions were excluded - both datasets in their entirety were used in training and evaluation. This means that for sequences with just one click, we require the trained embeddings to accurately describe the item, and time of viewing by the user to accurately classify the session, while for longer sessions, we can rely more on the RNN model to extract information from the sequence. This decision makes the training task much harder for our RNN model, but is a fairer comparison to previous work using GBM where all session lengths were also included \cite{Romov:2015:RCE:2813448.2813510},\cite{Volkovs:2015:TAI:2813448.2813512},\cite{Sheil2017}.

The RecSys 2015 Challenge dataset includes a dedicated test set, while the Retail Rocket dataset does not. We reserved 20\% of the Retail Rocket dataset for use in prediction / evaluation - the same proportion as the RecSys 2015 Challenge test dataset.

\begin{table}[!htb]
\centering
\begin{tabular}{ |c|c|c| } 
 \hline
  & RecSys 2015 & Retail Rocket \\
 \hline
 Sessions & \num{9249729} & \num{1398795} \\ 
 \hline
 Buyer sessions & 5.5\% & 0.7\% \\ 
 \hline
 Unique items & \num{52739} & \num{227006} \\ 
 \hline
\end{tabular}
\captionof{table}{A short comparison of the two datasets used - RecSys 2015 and Retail Rocket.}
\label{table:ds-compare}
\end{table}

\subsection{Data preparation}

The RecSys 2015 challenge dataset consists of 9.2 million user-item click sessions. Sessions are anonymous and classes are imbalanced with only 5\% of sessions ending in one or more buy events. Each user session captures the interactions between a single user and items or products : $S_n = {e_1,e_2,..,e_k}$, where $e_k$ is either a click or buy event. An example 2-event session is:

\begin{table}[!htb]
\centering
\begin{tabular}{ |c|c|c|c| } 
 \hline
 SID & Timestamp & Item ID & Cat ID \\ 
 \hline
 1 & 2014-04-07T10:51:09.277Z & 214536502 & 0 \\ 
 \hline
 1 & 2014-04-07T10:57:09.868Z & 214536500 & 0 \\
 \hline
\end{tabular}
\captionof{table}{An example of a clicker session from the RecSys 2015 dataset.}
\end{table}

Both datasets contain missing or obfuscated data - presumably for commercially sensitive reasons. Where sessions end with one or more purchase events, the item price and quantity values are provided only 30\% of the time in the case of the RecSys 2015 dataset, while prices are obfuscated for commercial reasons in the Retail Rocket dataset. Therefore these elements of the data provide limited value.

\begin{table}[!htb]
\centering
\begin{tabular}{ |c|c|c|c| }
 \hline
 SID & Item ID & Price & Quantity \\ 
 \hline
420374 & 214537888 & 12462 & 1 \\
 \hline
420374 & 214537850 & 10471 & 1 \\
 \hline
\end{tabular}
\captionof{table}{An example of the buy events from a buyer session (timestamp column elided for brevity).}
\end{table}

The Retail Rocket dataset consists of 1.4 million sessions. Sessions are also anonymous and are even more imbalanced - just 0.7\% of the sessions end in a buy event. This dataset also provides item metadata but in order to standardize our approach across both datasets, we chose not to use any information that was not common to both datasets. In particular we discard and do not use the additional "addtobasket" event that is present in the Retail Rocket dataset. Since it is so closely correlated with the buy event (users add to a basket before purchasing that basket), it renders the buyer prediction task trivial and an AUC of 0.97 is easily achievable for both our RNN and GBM models.

Our approach in preparing the data for training is as follows. We process each column as follows:
\begin{itemize}
  \item Session IDs are discarded (of course we retain the sequence grouping indicated by the IDs).
  \item Timestamps are quantized into bins 4 hours in duration.
  \item Item IDs are unchanged.
  \item Category IDs are unchanged.
  \item Purchase prices are unchanged. We calculate price variance per item to convey price movements to our model (e.g. a merchant special offer).
  \item Purchase quantities are unchanged.
\end{itemize}

Each field is then converted to an embedding vocabulary - simply a lookup table mapping values to integer IDs. We do not impose a minimum occurrence limit on any field - a value occurring even once will be represented in the respective embedding. This ensures that even "long tail" items will be presented to the model during training.
Lookup tables are then converted to an embedding with embedding weights initialized from a range \{-0.075, +0.075\} - Table \ref{table:emb} identifies the number of unique items per embedding and the width used. The testing dataset contains both item IDs and category IDs that are not present in the training set - however only a very small number of sessions are affected by this data in the test set.

This approach, combined with the use of Artificial Neural Networks, provides a learnable capacity to encode more information than just the original numeric value. For example, an item price of \$100 vs \$150 is not simply a numeric price difference, it can also signify learnable information on brand, premium vs value and so on.

\begin{table}[!htb]
\centering
\begin{tabular}{ |c|c|c|c| } 
 \hline
 Data name & Train & Train+Test & Embedding \\
 &&&Width \\ 
 \hline
 Item ID & \num{52739} & \num{54287} & 100 \\ 
 \hline
 Category ID & 340 & 348 & 10 \\ 
 \hline
 Timestamp & \num{4368} & \num{4368} & 10 \\ 
 \hline
 Price & 667 & 667 & 10 \\ 
 \hline
 Quantity & 1 & 1 & 10 \\ 
 \hline
\end{tabular}
\captionof{table}{Data field embeddings and dimensions, along with unique value counts for the training and test splits of the RecSys 2015 dataset.}
\label{table:emb}
\end{table}

\subsection{Event Unrolling}

In \cite{Bogina2017IncorporatingDT}, a more explicit representation of user dwelltime or interest in a particular item $i_k$ in a sequence $e_{i_1}, \ldots, e_{i_k}$ is provided to the model by repeating the presentation of the event containing the item to the model in proportion to the elapsed time between $e_{i_k}$ and $e_{i_{k+1}}$. In the example 2-event session displayed previously, the elapsed time between the first and second event is 6 minutes, therefore the first event is replayed 3 times during training and inference ($\ceil*{360 / 150}$).
In contrast to \cite{Bogina2017IncorporatingDT}, we found that session unrolling provided a smaller improvement in model performance - for example on the RecSys 2015 dataset our best AUC increased from 0.837 to 0.839 when using the optimal unrolling value (which we discovered empirically using grid search) of 150 seconds. Unrolling also comes with a cost of increasing session length and thus training time - Table \ref{table:unrolling} demonstrates the effect of session unrolling on the size of the training / validation and test sets on both datasets.

\begin{table}
\centering
\begin{tabular}{ |c|c|c|c| } 
 \hline
 Dataset & Events before & Events after & \% increase \\ 
 \hline
 RecSys 2015 & \num{41255735} & \num{56059913} & 36\% \\ 
 \hline
 Retail Rocket & \num{2351354} & \num{11224267} & 377\% \\ 
 \hline
\end{tabular}
\captionof{table}{Effect of unrolling by dwelltime on the RecSys 2015 and Retail Rocket datasets. There is a clear difference in the mean / median session duration of each dataset.}
\label{table:unrolling}
\end{table}

\subsection{Sequence Reversal}

From \cite{Sheil2017}, we know that the most important item in a user session is the last item (followed by the first item). Intuitively, the last and first items a user browses in a session are most predictive of purchase intent. To capitalize on this, we reversed the sequence order for each session before presenting them as batches to the model. Re-ordering the sequences provided an increase in test AUC on the RecSys 2015 dataset of 0.005 - from 0.834 to 0.839.

\subsection{Model}

\subsubsection{Model Architecture}

The data embedding modules are concatenated and presented to a configurable number of RNN layers (typically 3), with a final linear layer combining the output of the hidden units from the last layer. A sigmoid function is then applied to calculate a confidence probability in class membership.
The model is trained by minimizing an unweighted binary cross entropy loss:
\begin{equation}
    \label{eqn:test}
l_n = - \left[ y_n \cdot \log x_n + (1 - y_n) \cdot \log (1 - x_n) \right]
\end{equation}
where:\\
$x_n$ is the output label value from the model [0..1]\\
$y_n$ is the target label value \{0, 1\}.

We conducted a grid search over the number of layers and layer size by RNN type, as indicated in Table \ref{table:gridsearch} below.

\begin{table*}[!htb]
\centering
\begin{tabular}{ |c|c|c|c|c|c|c|c|c|c| }
 \hline
 &  \multicolumn{3}{c}{RNN} & \multicolumn{3}{c}{GRU} & \multicolumn{3}{c|}{LSTM}\\
Layers	& 1	& 2	& 3	& 1	& 2	& 3	& 1	& 2	& 3 \\
 \hline
 \multicolumn{10}{|l|}{Layer size} \\
 \hline
64	& 0.72	& 0.81	& 0.81	& 0.741	& 0.832	& 0.833	& 0.735	& 0.832	& 0.831 \\
 \hline
128	& 0.72	& 0.80	& 0.80	& 0.755	& 0.833	& 0.833	& 0.729	& 0.834	& 0.834 \\
 \hline
256	& 0.71	& 0.80	& 0.80	& 0.732	& 0.834	& 0.834	& 0.724	& 0.834	& \textbf{0.839} \\
 \hline
512	& 0.69	& 0.80	& 0.77	& 0.746	& 0.832	& 0.833	& 0.759	& 0.835	& \textbf{0.839} \\
 \hline
\end{tabular}
\caption{Model grid search results for number and size of RNN layers by RNN type on the RecSys 2015 dataset. The State of the Art baseline for comparison is 0.853.}
\label{table:gridsearch}
\end{table*}

\begin{center}
\includegraphics[width=9cm,height=9cm,keepaspectratio]{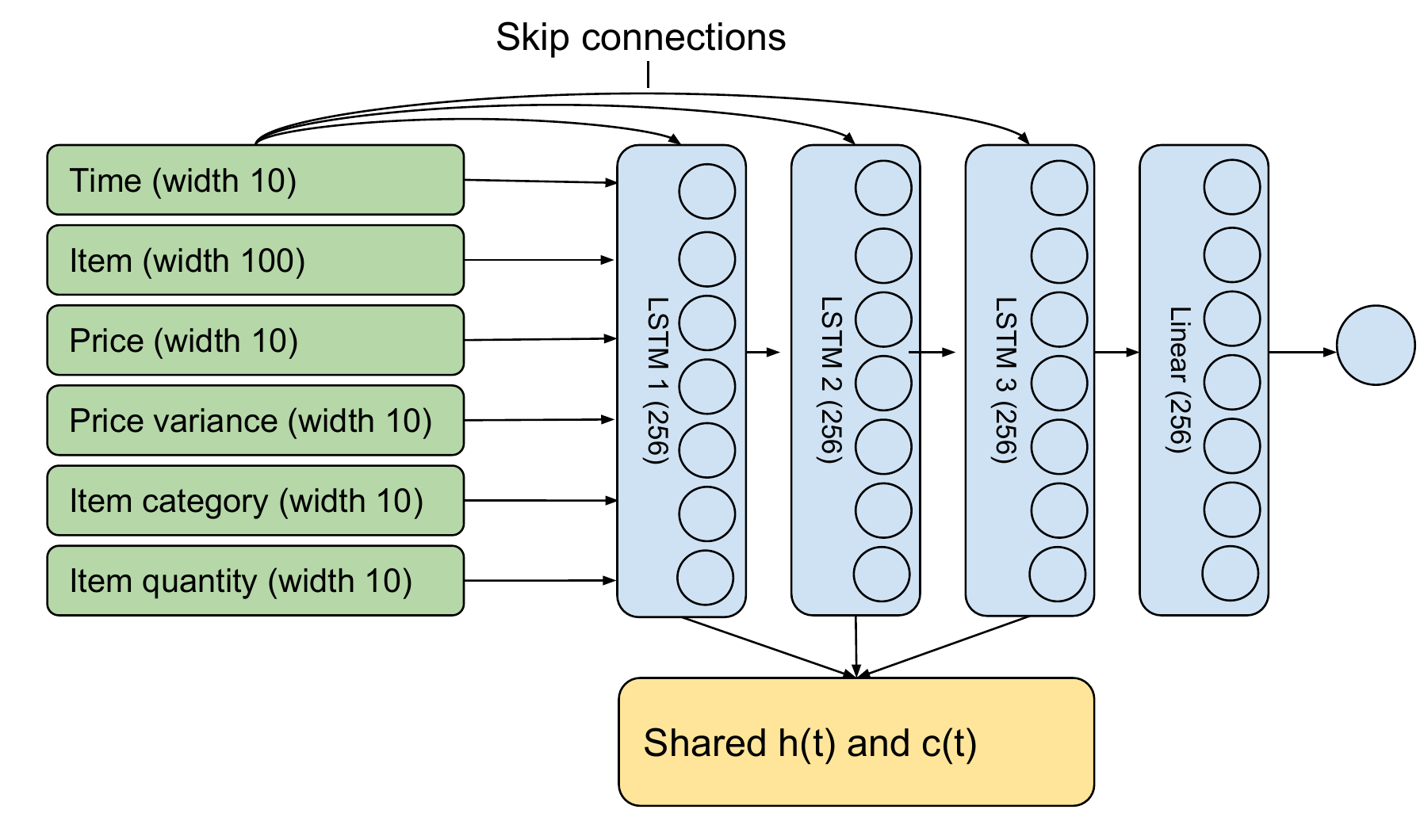}
\captionof{figure}{Model architecture used - the output is interpreted as the log probability that the input contains either a clicker or buyer session. Skip connections are used to combine the original input with successive layer outputs, and each layer shares the same hidden layer parameters.}
\end{center}

\subsubsection{Hidden layer parameter sharing}
One model design decision worthy of elaboration is how hidden information (and cell state for LSTM) is shared between the RNN layers. We found that best results were obtained by re-using hidden state across RNN layers - i.e. re-using the hidden state vector (and cell state vector for LSTM) from the previous layer in the architecture and initializing the next layer with these state vectors before presenting the next layer with the output from the previous layer. Our intuition here is that for Ecommerce datasets, the features learned by each layer are closely related due to the fact that Ecommerce log / clickstream data is very structured, therefore re-using hidden and cell states from lower layers helps higher layers to converge on important abstractions. Aggressive sharing of the hidden layers led to a very significant improvement in AUC, increasing from 0.75 to 0.84 in our single best model.

\section{Experiments and results}
In this section we describe the experimental setup, and the results obtained when comparing our best RNN model to GBM State of the Art, and a comparison of different RNN variants (vanilla, GRU, LSTM).

\subsection{Training Details}
Both datasets were split into a training set and validation set in a 90:10 ratio.  
The model was trained using the Adam optimizer \cite{DBLP:journals/corr/KingmaB14}, coupled with a binary cross entropy loss metric and a learning rate annealer. Training was halted after 2 successive epochs of worsening validation AUC. Table \ref{table:hyperparams} illustrates the main hyperparameters and setting used during training.

\begin{table}[ht]
\centering
\begin{tabular}{ |c|c| } 
 \hline
 Dataset split & $90/10$ (training / validation) \\ 
 \hline
 Hidden units range & $128 - 512: 256$ optimal \\ 
 \hline
 Embedding width  & $10 - 300: 100$ optimal for items\\ 
 \hline
 Embedding weight & $-0.075$ to $+0.075$\\ 
 \hline
 Batch size  & $32 - 256: 256$ optimal for speed \\
 & and regularization\\ 
 \hline
 Optimizer & Adam, learning rate $1e\textsuperscript{-3}$\\ 
 \hline
\end{tabular}
\caption{Hyper parameters and setup employed during model training.}
\label{table:hyperparams}
\end{table}

We tested three main types of recurrent cells (vanilla RNN, GRU, LSTM) as well as varying the number of cells per layer and layers per model. While a 3-layer LSTM achieved the best performance, vanilla RNNs which possess no memory mechanism are able to achieve a competitive AUC. The datasets are heavily weighted towards shorter session lengths (even after unrolling - see Figure \ref{fig:auc-by-sesslen} and \ref{fig:auc-by-sesslen-rr}). We posit that the power of LSTM and GRU is not needed for the shorter sequences of the dataset, and colloquial recurrence with embeddings has the capacity to model sessions over a short enough sequence length.

\subsection{Overall results}

The metric we used in our analysis was Area Under the ROC Curve or AUC. AUC is insensitive to class imbalance, and also the raw predictions from \cite{Romov:2015:RCE:2813448.2813510} were available to us, thus a detailed, like-for-like AUC comparison using the test set is the best model comparison. The organizers of the challenge also released the solution to the challenge, enabling the test set to be used.
After training, the LSTM model AUC obtained on the test set was 0.839 - ~98.4\% of the AUC (0.853) obtained by the SOTA model. As the subsequent experiments demonstrate, a combination of feature embeddings and model architecture decisions contribute to this performance.
For all model architecture variants tested (see Table \ref{table:gridsearch}), the best performance was achieved after training for a small number of epochs (2 - 3). This held true for both datasets.

Our LSTM model achieved within 98\% of the SOTA GBM performance on the RecSys 2015 dataset, and outperformed our GBM model by 0.5\% on the Retail Rocket dataset, as table \ref{table:model-compare} shows.

\begin{table}[!htb]
\centering
\begin{tabular}{ |c|c|c| } 
 \hline
  & RecSys 2015 & Retail Rocket \\
 \hline
 LSTM & \num{0.839} & \num[math-rm=\mathbf]{0.838} \\ 
 \hline
 GBM & \num[math-rm=\mathbf]{0.853} & \num{0.834} \\ 
 \hline
\end{tabular}
\captionof{table}{Classification performance measured using Area under the ROC curve (AUC) of the GBM and LSTM models on the RecSys 2015 and Retail Rocket datasets.}
\label{table:model-compare}
\end{table}

\begin{figure}
\centering
\includegraphics[width=8cm,height=8cm,keepaspectratio]{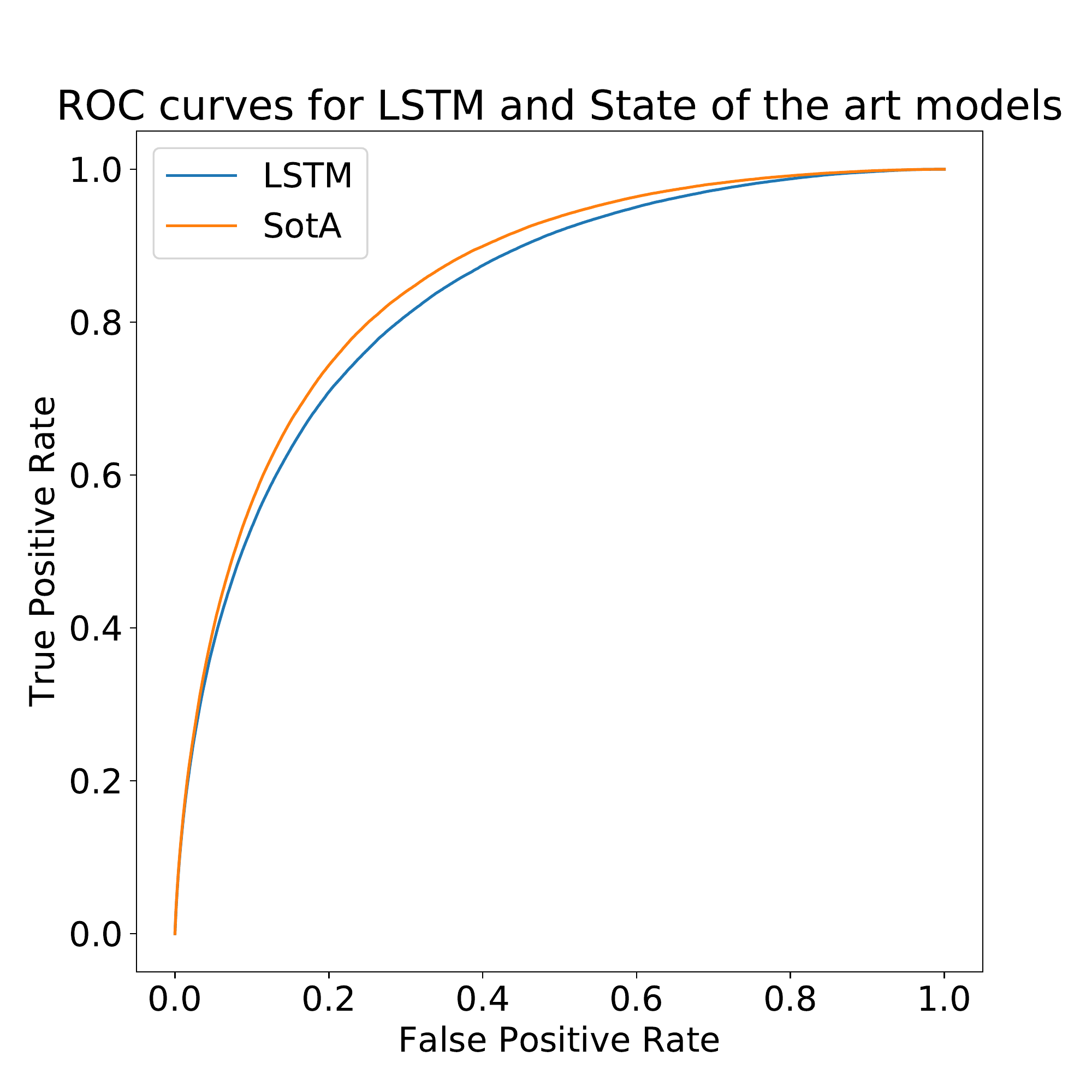}
\captionof{figure}{ROC curves for the LSTM and State of the Art models - on the RecSys 2015 test set.}
\end{figure}

\subsection{Analysis}

We constructed a number of tests to analyze model performance based on interesting subsets of the test data.

\subsubsection{Session length}
\begin{figure}
\centering
\includegraphics[width=9cm,height=9cm,keepaspectratio]{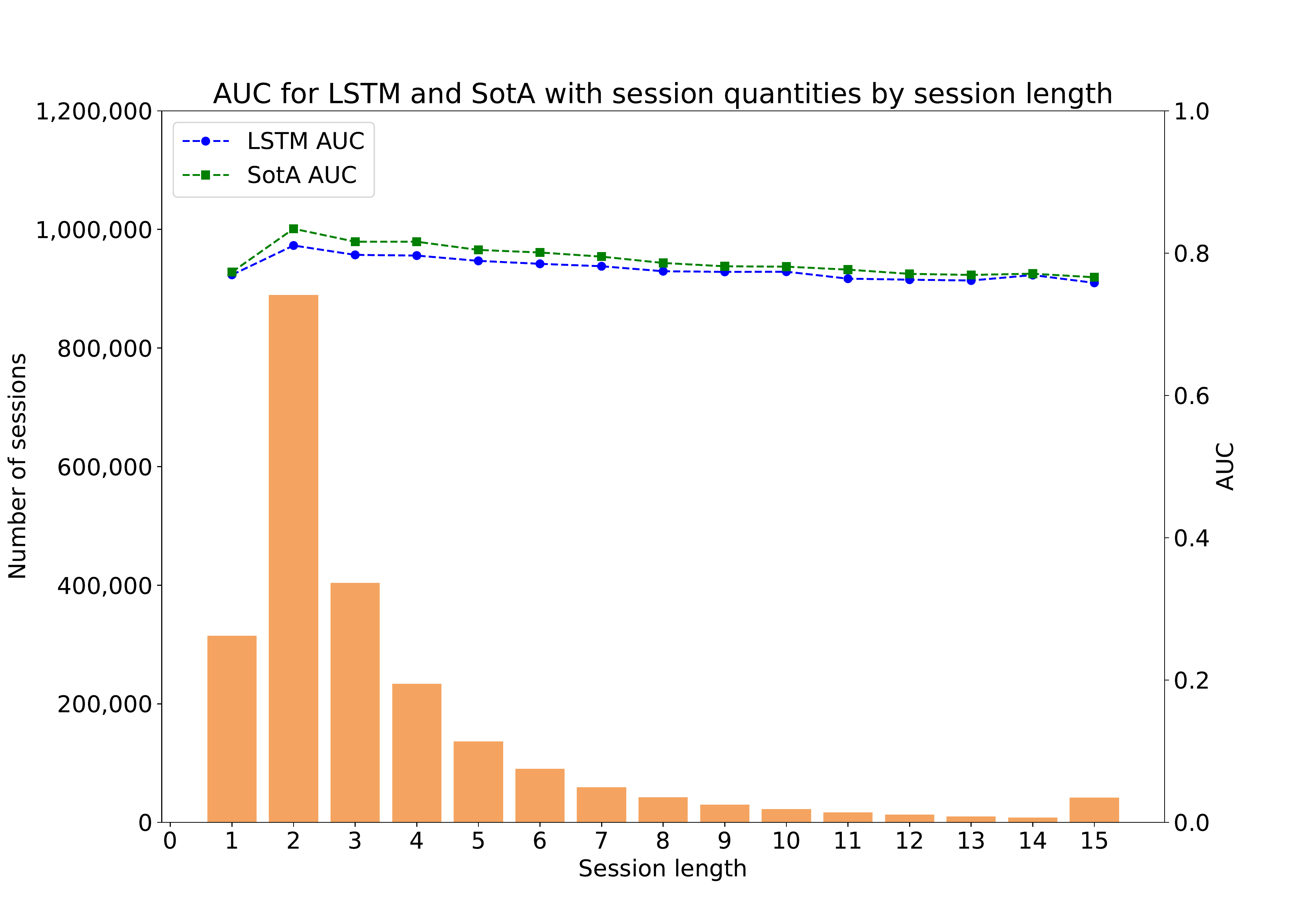}
\captionof{figure}{AUC by session length for the LSTM and SOTA models, session quantities by length also provided for context - clearly showing the bias towards short sequence / session lengths in the RecSys 2015 dataset.}\label{fig:auc-by-sesslen}
\end{figure}

Figure \ref{fig:auc-by-sesslen} graphs the best RNN model (a 3-layer LSTM with 256 cells per layer) and the SOTA model, with AUC scores broken down by session length. For context, the quantities for each session length in the test set is also provided. Both models underperform for sessions with just one click - clearly it is difficult to split clickers from buyers with such a small input signal. For the remaining session lengths, the relative model performance is consistent, although the LSTM model does start to close the gap after sessions with length $> 10$.

\subsubsection{User dwelltime}
\begin{figure}
\centering
\includegraphics[width=8cm,height=4cm,keepaspectratio]{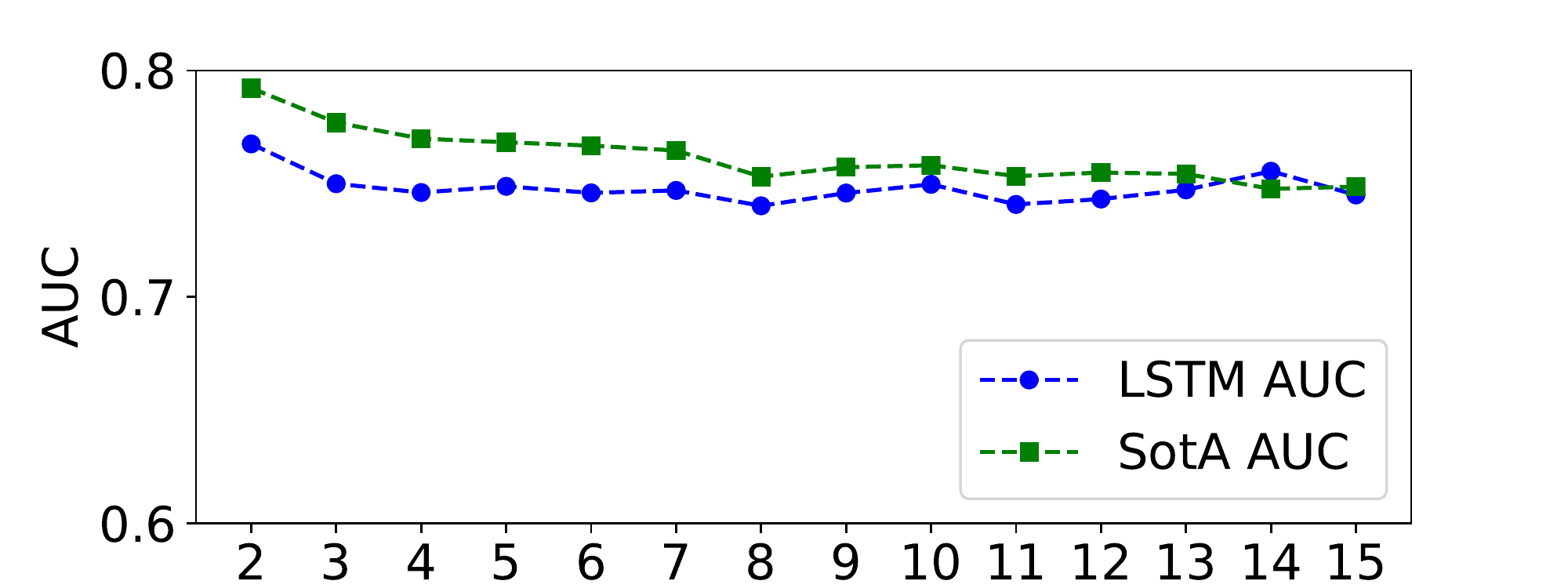}
\captionof{figure}{AUC by session length for the LSTM and SOTA models, for any sessions where unrolling by dwelltime was employed. There are no sessions of length 1 as unrolling is inapplicable for these sessions (RecSys 2015 dataset).}\label{fig:dwelltime}
\end{figure}

Given that we unrolled long-running events in order to provide more input to the RNN models, we evaluated the relative performance of each model when presented with sessions with any dwelltime $> 1$. As Figure \ref{fig:dwelltime} shows, LSTM is closer to SOTA for this subset of sessions and indeed outperforms SOTA for session length $= 14$, but the volume of sessions affected (\SI{5433}) is not enough to materially impact the overall AUC.

\subsubsection{Item price}
\begin{figure}[!htb]
\centering
\includegraphics[width=8cm,height=4cm,keepaspectratio]{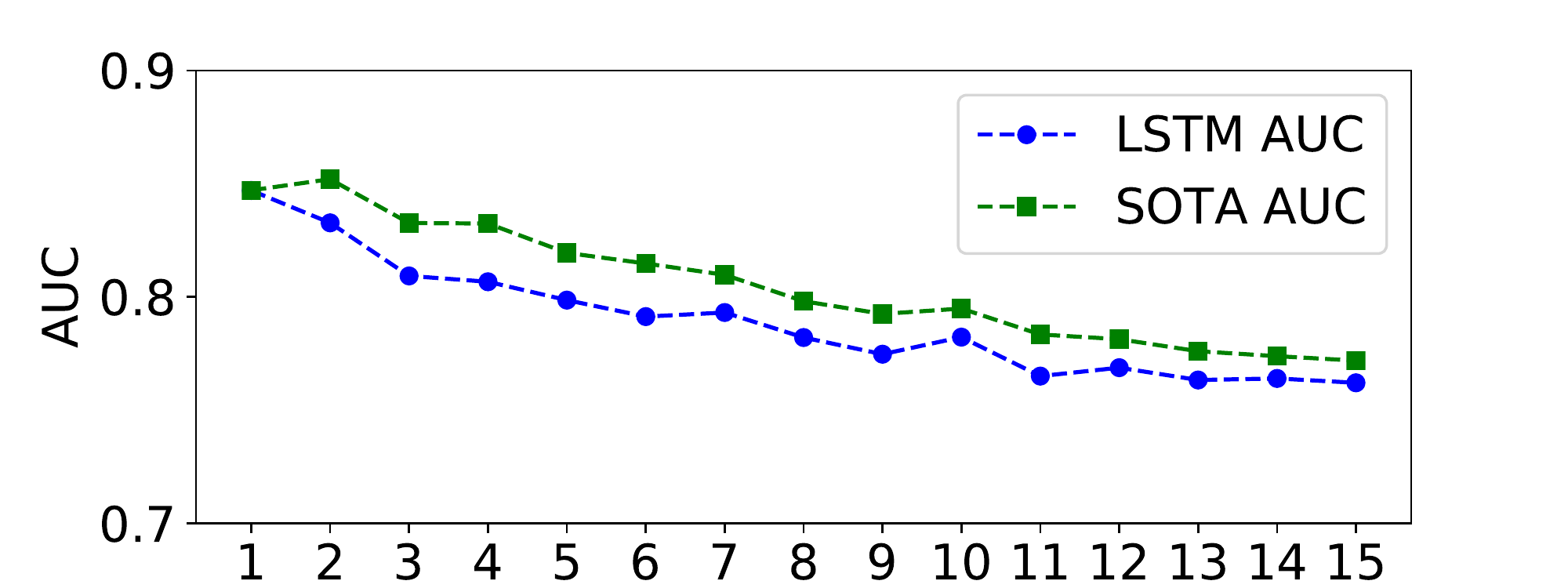}
\captionof{figure}{Model performance for sessions containing low price items, split by session length (RecSys 2015 dataset).}\label{fig:lowprice}
\end{figure}

\begin{figure}[!htb]
\centering
\includegraphics[width=8cm,height=4cm,keepaspectratio]{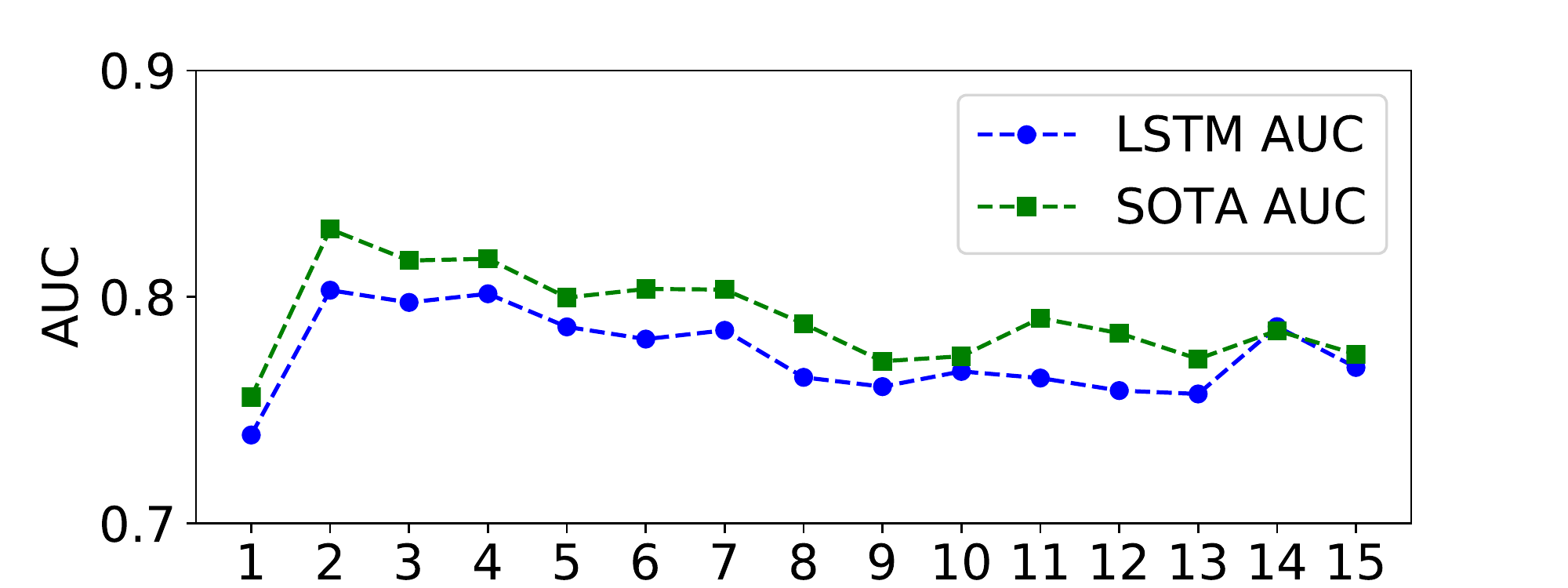}
\captionof{figure}{Model performance for sessions containing high price items, split by session length (RecSys 2015 dataset).}\label{fig:highprice}
\end{figure}

Like most Ecommerce catalogues, the catalogue under consideration here displays a considerable range of item prices. We first selected all sessions where any single item price was $>$ \SI{10000} (capturing \SI{544014} sessions) and then user sessions where the price was $<= 750$ (roughly 25\% of the maximum price - capturing \SI{1063034} sessions). Figures \ref{fig:lowprice} and \ref{fig:highprice} show the relative model performance for each session grouping. As with other selections, the relative model performance is broadly consistent - there is no region where LSTM either dramatically outperforms or underperforms the SOTA model. 


\subsubsection{Gated vs un-gated RNNs}

Table \ref{table:gridsearch} shows that while gated RNNs clearly outperform ungated or vanilla RNNs, the difference is 0.02 of AUC which is less than might be expected. We believe the reason for this is that the dataset contains many more shorter ($< 5$) than longer sequences. This is to be expected for anonymized data - user actions are only aggregated based on a current session token and there is no "lifetime" set of user events. For many real-world cases then, using ungated RNNs may deliver acceptable performance.

\subsubsection{End-to-end learning}

To measure the effect of allowing (or not) the gradients from the output loss to flow unencumbered throughout the model (including the embeddings), we froze the embedding layer so no gradient updates were applied and then trained the network. Model performance decreased to an AUC of 0.808 and training time increased by 3x to reach this AUC. Depriving the model of the ability to dynamically modify the input data representation  using gradients derived from the output loss metric reduces its ability to solve the classification problem posed.

\subsection{Transferring to another dataset}

\begin{figure}[!htb]
\centering
\includegraphics[width=8cm,height=8cm,keepaspectratio]{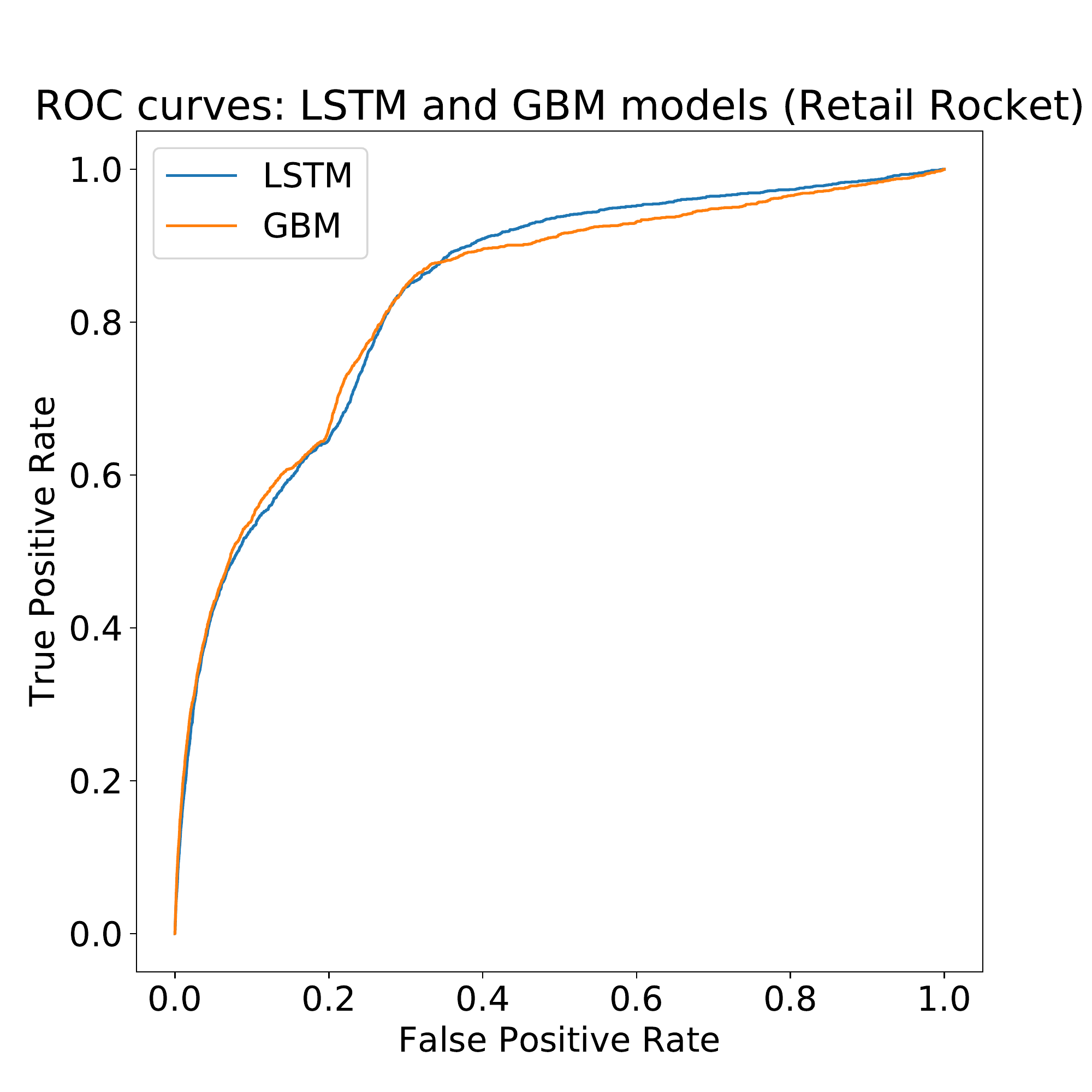}
\captionof{figure}{Area under ROC curves for LSTM and GBM models when ported to the Retail Rocket dataset. On this dataset, the LSTM model slightly outperforms the GBM model overall.}\label{fig:auc-rr}
\end{figure}

The GBM model itself used in \cite{Romov:2015:RCE:2813448.2813510} is not publically available, however we were able to use the GBM model described in \cite{Sheil2017}. Figure \ref{fig:auc-rr} shows the respective ROC curves for the RNN (LSTM) and GBM models when they are ported to the Retail Rocket dataset. Both models still perform well, however the LSTM model slightly out-performs the GBM model (AUC of 0.837 vs 0.834). 

\begin{figure}
\centering
\includegraphics[width=9cm,height=9cm,keepaspectratio]{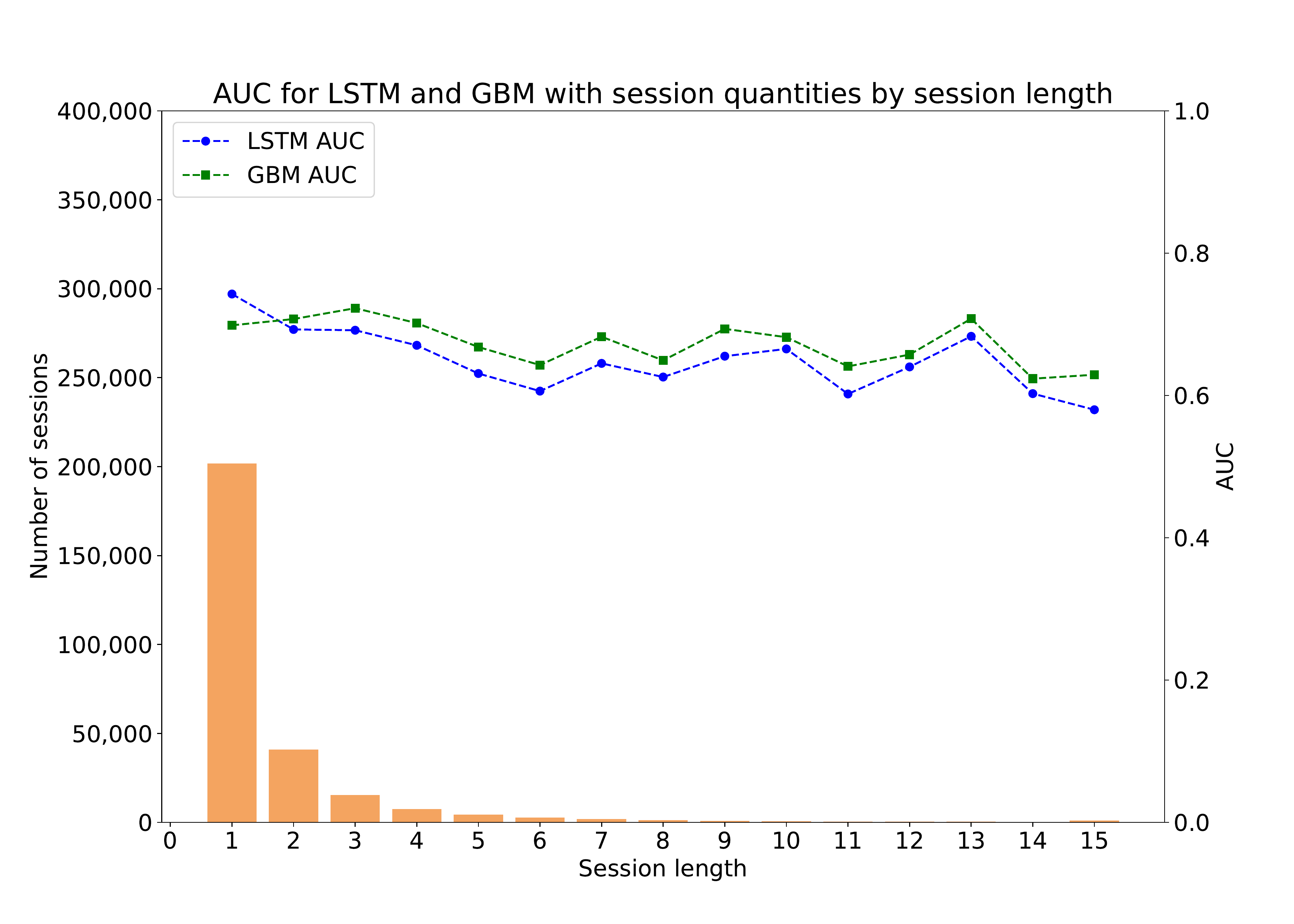}
\captionof{figure}{AUC by session length for the LSTM and GBM models when tested on the Retail Rocket dataset. The bias towards shorter sessions is even more prevalent versus the RecSys 2015 dataset.}
\label{fig:auc-by-sesslen-rr}
\end{figure}

A deeper analysis of the Area under the ROC curve demonstrates how the characteristics of the dataset can impact on model performance. The Retail Rocket dataset is heavily weighted towards single-click sessions as Figure \ref{fig:auc-by-sesslen-rr} shows. LSTM out-performs GBM for these sessions - which can be attributed more to the learned embeddings since there is no sequence to process. GBM by contrast can extract only limited value from single-click sessions as important feature components such as dwelltime, similarity etc. are unavailable.

\subsection{Training time and resources used}

We used PyTorch \cite{paszke2017automatic} to construct and train the LSTM models while XGBoost \cite{Chen:2016:XST:2939672.2939785} was used to train the GBM models. The LSTM implementation was trained on a single Nvidia GeForce GTX TITAN Black (circa 2014 and with single-precision performance of 5.1 TFLOPs vs a 2017 GTX 1080 Ti with 11.3 TFLOPs) and consumed between 1 and 2 GB RAM. The results reported were obtained after 3 epochs of training on the full dataset - an average of 6 hours (2 hours per epoch). This compares favourably to the training times and resources reported in \cite{Romov:2015:RCE:2813448.2813510}, where 150 machines were used for 12 hours. However, 12 hours denotes the time needed to train two models (session purchase and item ranking) whereas we train just one model. While we cannot compare GBM (which was trained on a CPU cluster) to RNN (trained on a single GPU with a different parallelization strategy) directly, we note that the hardware resources required for our model are modest and hence accessible to almost any commercial or academic setup. In addition, real-world Ecommerce datasets are large \cite{ebayarch} and change rapidly, therefore usable models must be able to consume large datasets and be re-trained readily to cater for new items / documents.

\subsection{Conclusions and future work}

We presented a Recurrent Neural Network (RNN) model which recovers 98.4\% of current SOTA performance on the user purchase prediction problem in Ecommerce without using explicit features. On a second dataset, our model fractionally exceeds SOTA performance. The model is straightforward to implement, generalizes to different datasets with comparable performance and can be trained with modest hardware resource requirements.

It is promising that gated RNNs with no feature engineering can be competitive with Gradient Boosted Machines on short session lengths and structured data - GBM is a  more established model choice in the domain of Recommender Systems and Ecommerce in general. We believe additional work on input representation (while still avoiding feature engineering) can further improve results for both gated and non-gated RNNs. One area of focus will be to investigate how parameter sharing at the hidden layer helps RNNs to operate on short sequences of structured data prevalent in Ecommerce.

Lastly,we note that although our approach requires no feature engineering, it is also inherently transductive - we plan to investigate embedding generation and maintenance approaches for new unseen items / documents to add an inductive capability to the architecture.

\section{Acknowledgements}

We would like to thank the authors of \cite{Romov:2015:RCE:2813448.2813510} for making their original test submission available and the organizers of the original challenge in releasing the solution file after the competition ended, enabling us to carry out our comparisons.

\bibliographystyle{ACM-Reference-Format}
\bibliography{sample-bibliography}


\begin{thebibliography}{37}


\ifx \showCODEN    \undefined \def \showCODEN     #1{\unskip}     \fi
\ifx \showDOI      \undefined \def \showDOI       #1{#1}\fi
\ifx \showISBNx    \undefined \def \showISBNx     #1{\unskip}     \fi
\ifx \showISBNxiii \undefined \def \showISBNxiii  #1{\unskip}     \fi
\ifx \showISSN     \undefined \def \showISSN      #1{\unskip}     \fi
\ifx \showLCCN     \undefined \def \showLCCN      #1{\unskip}     \fi
\ifx \shownote     \undefined \def \shownote      #1{#1}          \fi
\ifx \showarticletitle \undefined \def \showarticletitle #1{#1}   \fi
\ifx \showURL      \undefined \def \showURL       {\relax}        \fi
\providecommand\bibfield[2]{#2}
\providecommand\bibinfo[2]{#2}
\providecommand\natexlab[1]{#1}
\providecommand\showeprint[2][]{arXiv:#2}

\bibitem[\protect\citeauthoryear{Barkan and Koenigstein}{Barkan and
  Koenigstein}{2016}]%
        {DBLP:journals/corr/BarkanK16}
\bibfield{author}{\bibinfo{person}{Oren Barkan} {and} \bibinfo{person}{Noam
  Koenigstein}.} \bibinfo{year}{2016}\natexlab{}.
\newblock \showarticletitle{Item2Vec: Neural Item Embedding for Collaborative
  Filtering}.
\newblock \bibinfo{journal}{\emph{CoRR}}  \bibinfo{volume}{abs/1603.04259}
  (\bibinfo{year}{2016}).
\newblock
\showeprint[arxiv]{1603.04259}
\urldef\tempurl%
\url{http://arxiv.org/abs/1603.04259}
\showURL{%
\tempurl}


\bibitem[\protect\citeauthoryear{Ben-Shimon, Tsikinovsky, Friedmann, Shapira,
  Rokach, and Hoerle}{Ben-Shimon et~al\mbox{.}}{2015}]%
        {Ben-Shimon:2015:RCY:2792838.2798723}
\bibfield{author}{\bibinfo{person}{David Ben-Shimon},
  \bibinfo{person}{Alexander Tsikinovsky}, \bibinfo{person}{Michael Friedmann},
  \bibinfo{person}{Bracha Shapira}, \bibinfo{person}{Lior Rokach}, {and}
  \bibinfo{person}{Johannes Hoerle}.} \bibinfo{year}{2015}\natexlab{}.
\newblock \showarticletitle{RecSys Challenge 2015 and the YOOCHOOSE Dataset}.
  In \bibinfo{booktitle}{\emph{Proceedings of the 9th ACM Conference on
  Recommender Systems}} \emph{(\bibinfo{series}{RecSys '15})}.
  \bibinfo{publisher}{ACM}, \bibinfo{address}{New York, NY, USA},
  \bibinfo{pages}{357--358}.
\newblock
\showISBNx{978-1-4503-3692-5}
\urldef\tempurl%
\url{https://doi.org/10.1145/2792838.2798723}
\showDOI{\tempurl}


\bibitem[\protect\citeauthoryear{Bogina and Kuflik}{Bogina and Kuflik}{2017}]%
        {Bogina2017IncorporatingDT}
\bibfield{author}{\bibinfo{person}{Veronika Bogina} {and} \bibinfo{person}{Tsvi
  Kuflik}.} \bibinfo{year}{2017}\natexlab{}.
\newblock \showarticletitle{Incorporating Dwell Time in Session-Based
  Recommendations with Recurrent Neural Networks}. In
  \bibinfo{booktitle}{\emph{RecTemp@RecSys}}.
\newblock


\bibitem[\protect\citeauthoryear{Chen and Guestrin}{Chen and Guestrin}{2016}]%
        {Chen:2016:XST:2939672.2939785}
\bibfield{author}{\bibinfo{person}{Tianqi Chen} {and} \bibinfo{person}{Carlos
  Guestrin}.} \bibinfo{year}{2016}\natexlab{}.
\newblock \showarticletitle{XGBoost: A Scalable Tree Boosting System}. In
  \bibinfo{booktitle}{\emph{Proceedings of the 22Nd ACM SIGKDD International
  Conference on Knowledge Discovery and Data Mining}}
  \emph{(\bibinfo{series}{KDD '16})}. \bibinfo{publisher}{ACM},
  \bibinfo{address}{New York, NY, USA}, \bibinfo{pages}{785--794}.
\newblock
\showISBNx{978-1-4503-4232-2}
\urldef\tempurl%
\url{https://doi.org/10.1145/2939672.2939785}
\showDOI{\tempurl}


\bibitem[\protect\citeauthoryear{Cho, van Merrienboer, Bahdanau, and
  Bengio}{Cho et~al\mbox{.}}{2014}]%
        {DBLP:journals/corr/ChoMBB14}
\bibfield{author}{\bibinfo{person}{KyungHyun Cho}, \bibinfo{person}{Bart van
  Merrienboer}, \bibinfo{person}{Dzmitry Bahdanau}, {and}
  \bibinfo{person}{Yoshua Bengio}.} \bibinfo{year}{2014}\natexlab{}.
\newblock \showarticletitle{On the Properties of Neural Machine Translation:
  Encoder-Decoder Approaches}.
\newblock \bibinfo{journal}{\emph{CoRR}}  \bibinfo{volume}{abs/1409.1259}
  (\bibinfo{year}{2014}).
\newblock
\showeprint[arxiv]{1409.1259}
\urldef\tempurl%
\url{http://arxiv.org/abs/1409.1259}
\showURL{%
\tempurl}


\bibitem[\protect\citeauthoryear{Ding, Liu, Duan, and Nie}{Ding
  et~al\mbox{.}}{2015}]%
        {Ding:2015:MUC:2886521.2886653}
\bibfield{author}{\bibinfo{person}{Xiao Ding}, \bibinfo{person}{Ting Liu},
  \bibinfo{person}{Junwen Duan}, {and} \bibinfo{person}{Jian-Yun Nie}.}
  \bibinfo{year}{2015}\natexlab{}.
\newblock \showarticletitle{Mining User Consumption Intention from Social Media
  Using Domain Adaptive Convolutional Neural Network}. In
  \bibinfo{booktitle}{\emph{Proceedings of the Twenty-Ninth AAAI Conference on
  Artificial Intelligence}} \emph{(\bibinfo{series}{AAAI'15})}.
  \bibinfo{publisher}{AAAI Press}, \bibinfo{pages}{2389--2395}.
\newblock
\showISBNx{0-262-51129-0}
\urldef\tempurl%
\url{http://dl.acm.org/citation.cfm?id=2886521.2886653}
\showURL{%
\tempurl}


\bibitem[\protect\citeauthoryear{Friedman}{Friedman}{2000}]%
        {Friedman00greedyfunction}
\bibfield{author}{\bibinfo{person}{Jerome~H. Friedman}.}
  \bibinfo{year}{2000}\natexlab{}.
\newblock \showarticletitle{Greedy Function Approximation: A Gradient Boosting
  Machine}.
\newblock \bibinfo{journal}{\emph{Annals of Statistics}}  \bibinfo{volume}{29}
  (\bibinfo{year}{2000}), \bibinfo{pages}{1189--1232}.
\newblock


\bibitem[\protect\citeauthoryear{Goodfellow, Bengio, and Courville}{Goodfellow
  et~al\mbox{.}}{2016}]%
        {Goodfellow-et-al-2016}
\bibfield{author}{\bibinfo{person}{Ian Goodfellow}, \bibinfo{person}{Yoshua
  Bengio}, {and} \bibinfo{person}{Aaron Courville}.}
  \bibinfo{year}{2016}\natexlab{}.
\newblock \bibinfo{booktitle}{\emph{Deep Learning}}.
\newblock \bibinfo{publisher}{MIT Press}.
\newblock
\newblock
\shownote{\url{http://www.deeplearningbook.org}.}


\bibitem[\protect\citeauthoryear{Grbovic, Radosavljevic, Djuric, Bhamidipati,
  Savla, Bhagwan, and Sharp}{Grbovic et~al\mbox{.}}{2016}]%
        {DBLP:journals/corr/GrbovicRDBSBS16}
\bibfield{author}{\bibinfo{person}{Mihajlo Grbovic}, \bibinfo{person}{Vladan
  Radosavljevic}, \bibinfo{person}{Nemanja Djuric}, \bibinfo{person}{Narayan
  Bhamidipati}, \bibinfo{person}{Jaikit Savla}, \bibinfo{person}{Varun
  Bhagwan}, {and} \bibinfo{person}{Doug Sharp}.}
  \bibinfo{year}{2016}\natexlab{}.
\newblock \showarticletitle{E-commerce in Your Inbox: Product Recommendations
  at Scale}.
\newblock \bibinfo{journal}{\emph{CoRR}}  \bibinfo{volume}{abs/1606.07154}
  (\bibinfo{year}{2016}).
\newblock
\showeprint[arxiv]{1606.07154}
\urldef\tempurl%
\url{http://arxiv.org/abs/1606.07154}
\showURL{%
\tempurl}


\bibitem[\protect\citeauthoryear{Hidasi, Karatzoglou, Baltrunas, and
  Tikk}{Hidasi et~al\mbox{.}}{2015}]%
        {DBLP:journals/corr/HidasiKBT15}
\bibfield{author}{\bibinfo{person}{Bal{\'{a}}zs Hidasi},
  \bibinfo{person}{Alexandros Karatzoglou}, \bibinfo{person}{Linas Baltrunas},
  {and} \bibinfo{person}{Domonkos Tikk}.} \bibinfo{year}{2015}\natexlab{}.
\newblock \showarticletitle{Session-based Recommendations with Recurrent Neural
  Networks}.
\newblock \bibinfo{journal}{\emph{CoRR}}  \bibinfo{volume}{abs/1511.06939}
  (\bibinfo{year}{2015}).
\newblock
\showeprint[arxiv]{1511.06939}
\urldef\tempurl%
\url{http://arxiv.org/abs/1511.06939}
\showURL{%
\tempurl}


\bibitem[\protect\citeauthoryear{Hochreiter and Schmidhuber}{Hochreiter and
  Schmidhuber}{1997}]%
        {Hochreiter:1997:LSM:1246443.1246450}
\bibfield{author}{\bibinfo{person}{Sepp Hochreiter} {and}
  \bibinfo{person}{J\"{u}rgen Schmidhuber}.} \bibinfo{year}{1997}\natexlab{}.
\newblock \showarticletitle{Long Short-Term Memory}.
\newblock \bibinfo{journal}{\emph{Neural Comput.}} \bibinfo{volume}{9},
  \bibinfo{number}{8} (\bibinfo{date}{Nov.} \bibinfo{year}{1997}),
  \bibinfo{pages}{1735--1780}.
\newblock
\showISSN{0899-7667}
\urldef\tempurl%
\url{https://doi.org/10.1162/neco.1997.9.8.1735}
\showDOI{\tempurl}


\bibitem[\protect\citeauthoryear{Jannach, Ludewig, and Lerche}{Jannach
  et~al\mbox{.}}{2017}]%
        {Jannach2017SessionbasedIR}
\bibfield{author}{\bibinfo{person}{Dietmar Jannach}, \bibinfo{person}{Malte
  Ludewig}, {and} \bibinfo{person}{Lukas Lerche}.}
  \bibinfo{year}{2017}\natexlab{}.
\newblock \showarticletitle{Session-based item recommendation in e-commerce: on
  short-term intents, reminders, trends and discounts}.
\newblock \bibinfo{journal}{\emph{User Modeling and User-Adapted Interaction}}
  \bibinfo{volume}{27} (\bibinfo{year}{2017}), \bibinfo{pages}{351--392}.
\newblock


\bibitem[\protect\citeauthoryear{Karpathy, Johnson, and Li}{Karpathy
  et~al\mbox{.}}{2015}]%
        {DBLP:journals/corr/KarpathyJL15}
\bibfield{author}{\bibinfo{person}{Andrej Karpathy}, \bibinfo{person}{Justin
  Johnson}, {and} \bibinfo{person}{Fei{-}Fei Li}.}
  \bibinfo{year}{2015}\natexlab{}.
\newblock \showarticletitle{Visualizing and Understanding Recurrent Networks}.
\newblock \bibinfo{journal}{\emph{CoRR}}  \bibinfo{volume}{abs/1506.02078}
  (\bibinfo{year}{2015}).
\newblock
\showeprint[arxiv]{1506.02078}
\urldef\tempurl%
\url{http://arxiv.org/abs/1506.02078}
\showURL{%
\tempurl}


\bibitem[\protect\citeauthoryear{Kingma and Ba}{Kingma and Ba}{2014}]%
        {DBLP:journals/corr/KingmaB14}
\bibfield{author}{\bibinfo{person}{Diederik~P. Kingma} {and}
  \bibinfo{person}{Jimmy Ba}.} \bibinfo{year}{2014}\natexlab{}.
\newblock \showarticletitle{Adam: {A} Method for Stochastic Optimization}.
\newblock \bibinfo{journal}{\emph{CoRR}}  \bibinfo{volume}{abs/1412.6980}
  (\bibinfo{year}{2014}).
\newblock
\showeprint[arxiv]{1412.6980}
\urldef\tempurl%
\url{http://arxiv.org/abs/1412.6980}
\showURL{%
\tempurl}


\bibitem[\protect\citeauthoryear{Koren}{Koren}{2009}]%
        {Koren:2009:CFT:1557019.1557072}
\bibfield{author}{\bibinfo{person}{Yehuda Koren}.}
  \bibinfo{year}{2009}\natexlab{}.
\newblock \showarticletitle{Collaborative Filtering with Temporal Dynamics}. In
  \bibinfo{booktitle}{\emph{Proceedings of the 15th ACM SIGKDD International
  Conference on Knowledge Discovery and Data Mining}}
  \emph{(\bibinfo{series}{KDD '09})}. \bibinfo{publisher}{ACM},
  \bibinfo{address}{New York, NY, USA}, \bibinfo{pages}{447--456}.
\newblock
\showISBNx{978-1-60558-495-9}
\urldef\tempurl%
\url{https://doi.org/10.1145/1557019.1557072}
\showDOI{\tempurl}


\bibitem[\protect\citeauthoryear{Lipton}{Lipton}{2015}]%
        {DBLP:journals/corr/Lipton15}
\bibfield{author}{\bibinfo{person}{Zachary~Chase Lipton}.}
  \bibinfo{year}{2015}\natexlab{}.
\newblock \showarticletitle{A Critical Review of Recurrent Neural Networks for
  Sequence Learning}.
\newblock \bibinfo{journal}{\emph{CoRR}}  \bibinfo{volume}{abs/1506.00019}
  (\bibinfo{year}{2015}).
\newblock
\showeprint[arxiv]{1506.00019}
\urldef\tempurl%
\url{http://arxiv.org/abs/1506.00019}
\showURL{%
\tempurl}


\bibitem[\protect\citeauthoryear{Ludewig and Jannach}{Ludewig and
  Jannach}{2018}]%
        {DBLP:journals/corr/abs-1803-09587}
\bibfield{author}{\bibinfo{person}{Malte Ludewig} {and}
  \bibinfo{person}{Dietmar Jannach}.} \bibinfo{year}{2018}\natexlab{}.
\newblock \showarticletitle{Evaluation of Session-based Recommendation
  Algorithms}.
\newblock \bibinfo{journal}{\emph{CoRR}}  \bibinfo{volume}{abs/1803.09587}
  (\bibinfo{year}{2018}).
\newblock
\showeprint[arxiv]{1803.09587}
\urldef\tempurl%
\url{http://arxiv.org/abs/1803.09587}
\showURL{%
\tempurl}


\bibitem[\protect\citeauthoryear{Mikolov, Chen, Corrado, and Dean}{Mikolov
  et~al\mbox{.}}{2013}]%
        {DBLP:journals/corr/abs-1301-3781}
\bibfield{author}{\bibinfo{person}{Tomas Mikolov}, \bibinfo{person}{Kai Chen},
  \bibinfo{person}{Greg Corrado}, {and} \bibinfo{person}{Jeffrey Dean}.}
  \bibinfo{year}{2013}\natexlab{}.
\newblock \showarticletitle{Efficient Estimation of Word Representations in
  Vector Space}.
\newblock \bibinfo{journal}{\emph{CoRR}}  \bibinfo{volume}{abs/1301.3781}
  (\bibinfo{year}{2013}).
\newblock
\showeprint[arxiv]{1301.3781}
\urldef\tempurl%
\url{http://arxiv.org/abs/1301.3781}
\showURL{%
\tempurl}


\bibitem[\protect\citeauthoryear{Pascanu, Mikolov, and Bengio}{Pascanu
  et~al\mbox{.}}{2013}]%
        {Pascanu:2013:DTR:3042817.3043083}
\bibfield{author}{\bibinfo{person}{Razvan Pascanu}, \bibinfo{person}{Tomas
  Mikolov}, {and} \bibinfo{person}{Yoshua Bengio}.}
  \bibinfo{year}{2013}\natexlab{}.
\newblock \showarticletitle{On the Difficulty of Training Recurrent Neural
  Networks}. In \bibinfo{booktitle}{\emph{Proceedings of the 30th International
  Conference on International Conference on Machine Learning - Volume 28}}
  \emph{(\bibinfo{series}{ICML'13})}. \bibinfo{publisher}{JMLR.org},
  \bibinfo{pages}{III--1310--III--1318}.
\newblock
\urldef\tempurl%
\url{http://dl.acm.org/citation.cfm?id=3042817.3043083}
\showURL{%
\tempurl}


\bibitem[\protect\citeauthoryear{Paszke, Gross, Chintala, Chanan, Yang, DeVito,
  Lin, Desmaison, Antiga, and Lerer}{Paszke et~al\mbox{.}}{2017}]%
        {paszke2017automatic}
\bibfield{author}{\bibinfo{person}{Adam Paszke}, \bibinfo{person}{Sam Gross},
  \bibinfo{person}{Soumith Chintala}, \bibinfo{person}{Gregory Chanan},
  \bibinfo{person}{Edward Yang}, \bibinfo{person}{Zachary DeVito},
  \bibinfo{person}{Zeming Lin}, \bibinfo{person}{Alban Desmaison},
  \bibinfo{person}{Luca Antiga}, {and} \bibinfo{person}{Adam Lerer}.}
  \bibinfo{year}{2017}\natexlab{}.
\newblock \showarticletitle{Automatic differentiation in PyTorch}.
\newblock  (\bibinfo{year}{2017}).
\newblock


\bibitem[\protect\citeauthoryear{Pennington, Socher, and Manning}{Pennington
  et~al\mbox{.}}{2014}]%
        {pennington2014glove}
\bibfield{author}{\bibinfo{person}{Jeffrey Pennington},
  \bibinfo{person}{Richard Socher}, {and} \bibinfo{person}{Christopher~D.
  Manning}.} \bibinfo{year}{2014}\natexlab{}.
\newblock \showarticletitle{GloVe: Global Vectors for Word Representation}. In
  \bibinfo{booktitle}{\emph{Empirical Methods in Natural Language Processing
  (EMNLP)}}. \bibinfo{pages}{1532--1543}.
\newblock
\urldef\tempurl%
\url{http://www.aclweb.org/anthology/D14-1162}
\showURL{%
\tempurl}


\bibitem[\protect\citeauthoryear{Phillips}{Phillips}{2005}]%
        {phillips2005pricing}
\bibfield{author}{\bibinfo{person}{R.L. Phillips}.}
  \bibinfo{year}{2005}\natexlab{}.
\newblock \bibinfo{booktitle}{\emph{Pricing and Revenue Optimization}}.
\newblock \bibinfo{publisher}{Stanford University Press}.
\newblock
\showISBNx{9780804746984}
\showLCCN{2005009126}
\urldef\tempurl%
\url{https://books.google.co.uk/books?id=bXsyO06qikEC}
\showURL{%
\tempurl}


\bibitem[\protect\citeauthoryear{Pryzant, joo Chung, and Jurafsky}{Pryzant
  et~al\mbox{.}}{2017}]%
        {Pryzant2017PredictingSF}
\bibfield{author}{\bibinfo{person}{Reid Pryzant}, \bibinfo{person}{Young joo
  Chung}, {and} \bibinfo{person}{Dan Jurafsky}.}
  \bibinfo{year}{2017}\natexlab{}.
\newblock \showarticletitle{Predicting Sales from the Language of Product
  Descriptions}. In \bibinfo{booktitle}{\emph{ACM SIGIR Forum}}. ACM.
\newblock


\bibitem[\protect\citeauthoryear{Rendle}{Rendle}{2010}]%
        {Rendle:2010:FM:1933307.1934620}
\bibfield{author}{\bibinfo{person}{Steffen Rendle}.}
  \bibinfo{year}{2010}\natexlab{}.
\newblock \showarticletitle{Factorization Machines}. In
  \bibinfo{booktitle}{\emph{Proceedings of the 2010 IEEE International
  Conference on Data Mining}} \emph{(\bibinfo{series}{ICDM '10})}.
  \bibinfo{publisher}{IEEE Computer Society}, \bibinfo{address}{Washington, DC,
  USA}, \bibinfo{pages}{995--1000}.
\newblock
\showISBNx{978-0-7695-4256-0}
\urldef\tempurl%
\url{https://doi.org/10.1109/ICDM.2010.127}
\showDOI{\tempurl}


\bibitem[\protect\citeauthoryear{Retailrocket}{Retailrocket}{2017}]%
        {retailrocket}
\bibfield{author}{\bibinfo{person}{Retailrocket}.}
  \bibinfo{year}{2017}\natexlab{}.
\newblock \bibinfo{title}{{Retailrocket recommender system dataset}}.
\newblock
  \bibinfo{howpublished}{\url{https://www.kaggle.com/retailrocket/ecommerce-dataset}}.
    (\bibinfo{year}{2017}).
\newblock
\newblock
\shownote{[Online; accessed 01-Feb-2018].}


\bibitem[\protect\citeauthoryear{Romov and Sokolov}{Romov and Sokolov}{2015}]%
        {Romov:2015:RCE:2813448.2813510}
\bibfield{author}{\bibinfo{person}{Peter Romov} {and} \bibinfo{person}{Evgeny
  Sokolov}.} \bibinfo{year}{2015}\natexlab{}.
\newblock \showarticletitle{RecSys Challenge 2015: Ensemble Learning with
  Categorical Features}. In \bibinfo{booktitle}{\emph{Proceedings of the 2015
  International ACM Recommender Systems Challenge}}
  \emph{(\bibinfo{series}{RecSys '15 Challenge})}. \bibinfo{publisher}{ACM},
  \bibinfo{address}{New York, NY, USA}, Article \bibinfo{articleno}{1},
  \bibinfo{numpages}{4}~pages.
\newblock
\showISBNx{978-1-4503-3665-9}
\urldef\tempurl%
\url{https://doi.org/10.1145/2813448.2813510}
\showDOI{\tempurl}


\bibitem[\protect\citeauthoryear{Rumelhart, Hinton, and Williams}{Rumelhart
  et~al\mbox{.}}{1986}]%
        {Rumelhart:1986:LIR:104279.104293}
\bibfield{author}{\bibinfo{person}{D.~E. Rumelhart}, \bibinfo{person}{G.~E.
  Hinton}, {and} \bibinfo{person}{R.~J. Williams}.}
  \bibinfo{year}{1986}\natexlab{}.
\newblock \showarticletitle{Parallel Distributed Processing: Explorations in
  the Microstructure of Cognition, Vol. 1}.
\newblock \bibinfo{publisher}{MIT Press}, \bibinfo{address}{Cambridge, MA,
  USA}, Chapter Learning Internal Representations by Error Propagation,
  \bibinfo{pages}{318--362}.
\newblock
\showISBNx{0-262-68053-X}
\urldef\tempurl%
\url{http://dl.acm.org/citation.cfm?id=104279.104293}
\showURL{%
\tempurl}


\bibitem[\protect\citeauthoryear{Sculley, Holt, Golovin, Davydov, Phillips,
  Ebner, Chaudhary, and Young}{Sculley et~al\mbox{.}}{2014}]%
        {high-interest}
\bibfield{author}{\bibinfo{person}{D. Sculley}, \bibinfo{person}{Gary Holt},
  \bibinfo{person}{Daniel Golovin}, \bibinfo{person}{Eugene Davydov},
  \bibinfo{person}{Todd Phillips}, \bibinfo{person}{Dietmar Ebner},
  \bibinfo{person}{Vinay Chaudhary}, {and} \bibinfo{person}{Michael Young}.}
  \bibinfo{year}{2014}\natexlab{}.
\newblock \showarticletitle{Machine Learning: The High Interest Credit Card of
  Technical Debt}. In \bibinfo{booktitle}{\emph{SE4ML: Software Engineering for
  Machine Learning (NIPS 2014 Workshop)}}.
\newblock


\bibitem[\protect\citeauthoryear{Sheil and Rana}{Sheil and Rana}{2017}]%
        {Sheil2017}
\bibfield{author}{\bibinfo{person}{Humphrey Sheil} {and} \bibinfo{person}{Omer
  Rana}.} \bibinfo{year}{2017}\natexlab{}.
\newblock \showarticletitle{Classifying and Recommending Using Gradient Boosted
  Machines and Vector Space Models}. In \bibinfo{booktitle}{\emph{Advances in
  Computational Intelligence Systems. UKCI 2017.}},
  \bibfield{editor}{\bibinfo{person}{Zhang~Q Chao~F., Schockaert~S.}} (Ed.),
  Vol.~\bibinfo{volume}{650}. \bibinfo{publisher}{Springer},
  \bibinfo{address}{Cham}.
\newblock
\urldef\tempurl%
\url{https://doi.org/10.1007/978-3-319-66939-7_18}
\showDOI{\tempurl}


\bibitem[\protect\citeauthoryear{Tan, Xu, and Liu}{Tan et~al\mbox{.}}{2016}]%
        {DBLP:journals/corr/TanXL16}
\bibfield{author}{\bibinfo{person}{Yong~Kiam Tan}, \bibinfo{person}{Xinxing
  Xu}, {and} \bibinfo{person}{Yong Liu}.} \bibinfo{year}{2016}\natexlab{}.
\newblock \showarticletitle{Improved Recurrent Neural Networks for
  Session-based Recommendations}.
\newblock \bibinfo{journal}{\emph{CoRR}}  \bibinfo{volume}{abs/1606.08117}
  (\bibinfo{year}{2016}).
\newblock
\showeprint[arxiv]{1606.08117}
\urldef\tempurl%
\url{http://arxiv.org/abs/1606.08117}
\showURL{%
\tempurl}


\bibitem[\protect\citeauthoryear{Toth, Tan, Di~Fabbrizio, and Datta}{Toth
  et~al\mbox{.}}{2017}]%
        {toth2017predicting}
\bibfield{author}{\bibinfo{person}{Arthur Toth}, \bibinfo{person}{Louis Tan},
  \bibinfo{person}{Giuseppe Di~Fabbrizio}, {and} \bibinfo{person}{Ankur
  Datta}.} \bibinfo{year}{2017}\natexlab{}.
\newblock \showarticletitle{Predicting Shopping Behavior with Mixture of RNNs}.
  In \bibinfo{booktitle}{\emph{ACM SIGIR Forum}}. ACM.
\newblock


\bibitem[\protect\citeauthoryear{Trotman, Degenhardt, and Kallumadi}{Trotman
  et~al\mbox{.}}{2017}]%
        {ebayarch}
\bibfield{author}{\bibinfo{person}{Andrew Trotman}, \bibinfo{person}{Jon
  Degenhardt}, {and} \bibinfo{person}{Surya Kallumadi}.}
  \bibinfo{year}{2017}\natexlab{}.
\newblock \showarticletitle{The Architecture of eBay Search}. In
  \bibinfo{booktitle}{\emph{ACM SIGIR Forum}}. ACM.
\newblock


\bibitem[\protect\citeauthoryear{Vezhnevets and Barinova}{Vezhnevets and
  Barinova}{2007}]%
        {Vezhnevets:2007:ABO:1421665.1421707}
\bibfield{author}{\bibinfo{person}{Alexander Vezhnevets} {and}
  \bibinfo{person}{Olga Barinova}.} \bibinfo{year}{2007}\natexlab{}.
\newblock \showarticletitle{Avoiding Boosting Overfitting by Removing Confusing
  Samples}. In \bibinfo{booktitle}{\emph{Proceedings of the 18th European
  Conference on Machine Learning}} \emph{(\bibinfo{series}{ECML '07})}.
  \bibinfo{publisher}{Springer-Verlag}, \bibinfo{address}{Berlin, Heidelberg},
  \bibinfo{pages}{430--441}.
\newblock
\showISBNx{978-3-540-74957-8}
\urldef\tempurl%
\url{https://doi.org/10.1007/978-3-540-74958-5_40}
\showDOI{\tempurl}


\bibitem[\protect\citeauthoryear{Volkovs}{Volkovs}{2015}]%
        {Volkovs:2015:TAI:2813448.2813512}
\bibfield{author}{\bibinfo{person}{Maksims Volkovs}.}
  \bibinfo{year}{2015}\natexlab{}.
\newblock \showarticletitle{Two-Stage Approach to Item Recommendation from User
  Sessions}. In \bibinfo{booktitle}{\emph{Proceedings of the 2015 International
  ACM Recommender Systems Challenge}} \emph{(\bibinfo{series}{RecSys '15
  Challenge})}. \bibinfo{publisher}{ACM}, \bibinfo{address}{New York, NY, USA},
  Article \bibinfo{articleno}{3}, \bibinfo{numpages}{4}~pages.
\newblock
\showISBNx{978-1-4503-3665-9}
\urldef\tempurl%
\url{https://doi.org/10.1145/2813448.2813512}
\showDOI{\tempurl}


\bibitem[\protect\citeauthoryear{Wu, Ahmed, Beutel, Smola, and Jing}{Wu
  et~al\mbox{.}}{2017}]%
        {Wu:2017:RRN:3018661.3018689}
\bibfield{author}{\bibinfo{person}{Chao-Yuan Wu}, \bibinfo{person}{Amr Ahmed},
  \bibinfo{person}{Alex Beutel}, \bibinfo{person}{Alexander~J. Smola}, {and}
  \bibinfo{person}{How Jing}.} \bibinfo{year}{2017}\natexlab{}.
\newblock \showarticletitle{Recurrent Recommender Networks}. In
  \bibinfo{booktitle}{\emph{Proceedings of the Tenth ACM International
  Conference on Web Search and Data Mining}} \emph{(\bibinfo{series}{WSDM
  '17})}. \bibinfo{publisher}{ACM}, \bibinfo{address}{New York, NY, USA},
  \bibinfo{pages}{495--503}.
\newblock
\showISBNx{978-1-4503-4675-7}
\urldef\tempurl%
\url{https://doi.org/10.1145/3018661.3018689}
\showDOI{\tempurl}


\bibitem[\protect\citeauthoryear{Yan, Zhou, and Duan}{Yan
  et~al\mbox{.}}{2015}]%
        {Yan:2015:EIR:2813448.2813511}
\bibfield{author}{\bibinfo{person}{Peng Yan}, \bibinfo{person}{Xiaocong Zhou},
  {and} \bibinfo{person}{Yitao Duan}.} \bibinfo{year}{2015}\natexlab{}.
\newblock \showarticletitle{E-Commerce Item Recommendation Based on Field-aware
  Factorization Machine}. In \bibinfo{booktitle}{\emph{Proceedings of the 2015
  International ACM Recommender Systems Challenge}}
  \emph{(\bibinfo{series}{RecSys '15 Challenge})}. \bibinfo{publisher}{ACM},
  \bibinfo{address}{New York, NY, USA}, Article \bibinfo{articleno}{2},
  \bibinfo{numpages}{4}~pages.
\newblock
\showISBNx{978-1-4503-3665-9}
\urldef\tempurl%
\url{https://doi.org/10.1145/2813448.2813511}
\showDOI{\tempurl}


\bibitem[\protect\citeauthoryear{Zhu, Li, Liao, Wang, Guan, Liu, and Cai}{Zhu
  et~al\mbox{.}}{2017}]%
        {Zhu2017WhatTD}
\bibfield{author}{\bibinfo{person}{Yu Zhu}, \bibinfo{person}{Hao Li},
  \bibinfo{person}{Yikang Liao}, \bibinfo{person}{Beidou Wang},
  \bibinfo{person}{Ziyu Guan}, \bibinfo{person}{Haifeng Liu}, {and}
  \bibinfo{person}{Deng Cai}.} \bibinfo{year}{2017}\natexlab{}.
\newblock \showarticletitle{What to Do Next: Modeling User Behaviors by
  Time-LSTM}. In \bibinfo{booktitle}{\emph{IJCAI}}.
\newblock


\end{thebibliography}

\end{document}